\pdfoutput=1
\documentclass[11pt]{article}

\usepackage{acl}

\usepackage{times}
\usepackage{latexsym}

\usepackage[T1]{fontenc}
\usepackage[utf8]{inputenc}

\usepackage{microtype}

\usepackage{inconsolata}

\usepackage{natbib}
\usepackage{multibib}
\usepackage{graphicx}
\usepackage{tabularx}
\usepackage{soul}
\usepackage{titlesec}

\usepackage{xcolor}
\usepackage{hyperref}
 \definecolor{darkblue}{rgb}{0, 0, 0.5}
  \hypersetup{colorlinks=true, citecolor=darkblue, linkcolor=darkblue, urlcolor=darkblue}

\usepackage{xstring}
\usepackage{color}

\usepackage{capt-of}
\usepackage{blindtext}

\usepackage{paralist}

\usepackage{graphicx}
\usepackage{amsmath}

\usepackage[normalem]{ulem}
\useunder{\uline}{\ul}{}

\usepackage[para]{footmisc}

\usepackage{times}
\usepackage{latexsym}
\usepackage{paralist}

\usepackage{comment}
\usepackage{booktabs}
\usepackage{amsmath,amssymb}
\usepackage{caption}
\usepackage{subcaption}
\usepackage{soul}
\usepackage{xcolor}

\usepackage{array,multirow,graphicx}

\usepackage{dirtytalk}

\title{Analyzing Semantic Change  through Lexical Replacements}

\author{Francesco Periti* \ \ \\
  {\small University of Milan} \ \ \\
  {\small Via Celoria 18,} \ \ \\
   {\small 20133, Milan, Italy} \ \ \\
  {\small \texttt{francesco.periti@unimi.it} \ \ } \\\And
  \ Pierluigi Cassotti* \ \ \\
  {\small University of Gothenburg} \ \ \\
  {\small Renstr\"{o}msgatan 6} \ \ \\
  {\small 40530 Gothenburg, Sweden} \ \ \\
  {\small \texttt{pierluigi.cassotti@gu.se} \ \ } \\\And
 Haim Dubossarsky \ \ \\
  {\small Queen Mary University of London} \ \ \\
  {\small Mile End Road} \ \ \\
  {\small E1 4NS London, England} \ \ \\
  {\small \texttt{h.dubossarsky@qmul.ac.uk} \ \ } \\\And
  Nina Tahmasebi \ \ \\
  {\small University of Gothenburg} \ \ \\
  {\small Renstr\"{o}msgatan 6} \ \ \\
  {\small 40530 Gothenburg, Sweden} \ \ \\
  {\small \texttt{nina.tahmasebi@gu.se} \ \ }\\
  }

\begin{document}
\maketitle
\begingroup\def\thefootnote{*}\footnotetext{These authors contributed equally}\endgroup

\begin{abstract}
Modern language models are capable of contextualizing words based on their surrounding context. However, this capability is often compromised due to semantic change that leads to words being used in new, unexpected contexts not encountered during pre-training. In this paper, we model \textit{semantic change} by studying the effect of unexpected contexts introduced by \textit{lexical replacements}. We propose a \textit{replacement schema} where a target word is substituted with lexical replacements of varying relatedness, thus simulating different kinds of semantic change. Furthermore, we leverage the replacement schema as a basis for a novel \textit{interpretable} model for semantic change. We are also the first to evaluate the use of LLaMa for semantic change detection. 

%1. Luigi needs to read contributions, lexical substitution and conclusions (P)
%put substitutes in Appendix 
%read limitations (P+F)

%PUT CORRECT scores in 4.1
\end{abstract}

\section{Introduction}\label{sec:introduction}
The major advancement that novel Language Models (LMs) have brought is the ability to dynamically generate contextualized representations  (i.e., embeddings) based on specific usage context. When words are used in contexts similar to those encountered during training, LMs can %thus 
easily differentiate, in a computational way, between word meanings. Like in the case of  \textit{rock} in the sentences \textit{sitting on a rock} and \textit{listening to rock}. 

However, when an existing word in our vocabulary gains a new meaning through semantic change, LMs' ability to differentiate that meaning can be affected. This stems from the fact that semantic change is evidenced through new contexts that were previously unknown for the word. Sometimes, the new meaning is novel to the dictionary, for example, the metaphorical Web-meaning of \textit{surfing}. Other times, the meaning is already in existence and get the word as a new referent. This is, for example, the case for \textit{happy}. It used to mean exclusively \texttt{to be lucky} and then gained the meaning of \texttt{happiness}. %Francesco: I think we need a reference here
In an inverse process, the word \textit{gay} lost its meaning of \texttt{happiness} and began to refer exclusively to \texttt{homosexuality}. One can think of this process of \textit{semantic change} to be a \textit{lexical replacement} of the word \textit{happy} into the context of \textit{gay}, like in the following sentence 

\begin{center}
The heart is sportive, light, and \textbf{gay}, \\life seems a long glad summer's day\footnote{Manchester Times, Wednesday 03 May 1854, found via \url{https://discovery.nationalarchives.gov.uk}.}
\end{center}

When using LMs, the representation of a word $w$ is based on 
(i) the pre-trained knowledge that the model has about $w$ given its position in the context, and (ii) the context $c$ in which $w$ is used. Thus, when this replacement happens, LMs experience a \textit{tension} between the {\color{blue}existing} sense/s of \textit{happy} (which do not include \texttt{happiness}) and the meaning of the {\color{red} new} context (which does indicate \texttt{happiness}). Due to semantic change, LMs do not known the relationship between the new context $c$ and the replacement word $r$. As a consequence, the representation of $r$ (i.e., \textit{happy} in the sense \texttt{to be lucky}) and the representation of $c$ (i.e., the context of \textit{gay} in the sense of \texttt{happiness}) pull in different directions challenging the LMs' ability to contextualize~\cite{ethayarajh2019contextual}.    

The  tension
%Nina: I removed the empahsis on pull as we discuss tension later in the paper
increases as the  gap between the data used for training the model, and the data on which the model is applied grows larger. Indeed, the LMs we use serve as the lens through which we view the studied texts: if our texts are contemporary with the pre-training, the gap is likely to be minimal. If, however, we intend to study historical or other out-of-domain  corpora through LMs trained on modern text, this gap can be arbitrarily large and have major effects on follow-up studies. Thus, using LMs for modeling relationships beyond their pre-trained knowledge will likely result in an
underestimation of semantic change. %Francesco: let's do a final check later

\paragraph{Our contributions:}%Francesco: I would like to have this section in a new page
In this paper, we propose a replacement schema to study the tension experienced by LMs when words undergo semantic change. Such schema involves replacing a word $w$ in the context $c$ with a replacement $r$ to analyze how the representation of $r$ differs from the original representation of $w$.

%paragraph for lms
Given a word $w$, our experiments systematically show that LMs (i.e., BERT, mBERT, XLM-R) experience a tension between the pre-trained knowledge of $w$ and the new context of a gained meaning. This tension  differs across linguistic relations, namely synonymy, antonymy, and hypernymy. 

%paragrapg for change
We then use the introduced schema for detecting semantic change. Our experiments show that, when random replacements are used to simulate \textit{synthetic} semantic change, the use of a clustering algorithm (i.e, Affinity Propagation) falls short to differentiate meanings and detect such change. Furthermore, we use the replacement schema to introduce a new \textit{interpretable} model for semantic change detection, while being comparable with state-of-the-art for English.

%paragraph for substitutions
%\textbf{TODO}
Finally, to further investigate the tension of LMs and semantic change, we compare the use of a pre-defined set of \textit{replacements} with word \textit{substitutes} generated by LMs (i.e., BERT, LLaMa 2). Our experiments show that smaller LMs are less able to provide substitutes that handle changing contexts, while LLaMa 2 significantly outperform the LMs. Notably, to the best of our knowledge, this represents the first experiment in the current literature to use LLaMa 2 to model semantic change and only the third paper to use any generative model \cite{periti2024chatgpt,periti2024systematic}. 

\section{Related Work}\label{sec:related-work}
For this paper, relevant work pertains both to  contextualization  of modern LMs and the field of lexical semantic change. 

\paragraph{Modern \textit{contextualized} LMs} leverage the Transformer architecture to capture the semantics of words~\cite{vaswani2017attention}. Their success in solving NLP tasks has prompted numerous studies to explore the nature and characteristics of their \textit{contextualization} ability. \newcite{ethayarajh2019contextual,coenen2019visualizing,cai2021isotropy,jawahar2019bert} shed light on the geometry of the embedding space. \newcite{serrano2019attention,bai2021attentions,guan2020far} investigate the interpretability of the attention mechanism. \newcite{yenicelik2020bert,gari2021lets,kalinowski2021exploring,haber2021patterns} examine the clusterability of word representations. \newcite{abdou2022word,hessel2021effective,mickus2020mean,wang2021on} analyze the impact of word position in the embeddings generation. \newcite{coenen2019visualizing,levine2020sensebert,pedinotti2020dont} study how word meaning are represented in the embedding space. 

%Most of the current work involves probing tasks, as proposed by~\citet{hewitt2019designing}. These tasks consist of training an auxiliary classifier on top of a model, where the contextualized embeddings serve as features to predict syntactic (e.g. PoS) and semantic (e.g. word relations) properties of words~\cite{clark2019bert,lin2022bert,Wallat2023probing,lin2022bert,ravichander2020systematicity}. If the auxiliary classifier accurately predicts a linguistic property, the property is assumed to be encoded in the model.  
Recent work have focused on a related aspect, namely adapting LMs to improve their \textit{temporal} contextualization. This challenge has been addressed across various applications such as named entity recognition~\cite{rijhwani2020temporally}, fake news detection~\cite{hu2023learn}, text summarization~\cite{cheang2023lms}, and lexical semantic change~\cite{su2022improving,rosin2022time,rosin2022temporal}. Nonetheless, while temporal domain adaptation can improve performance across various tasks,~\newcite{agarwal2022temporal} demonstrated that temporal contextualization %deterioration
may not always be a concern. In our work we complement existing research by using lexical replacements as a proxy to analyze how language models contextualize words that have undergone semantic change.

\paragraph{Lexical Semantic Change (LSC)} is the task of automatically identifying words that change their meaning over time~\cite{schlechtweg2020semeval}. The task is recently gaining more and more attention, acting as evaluation task to assess the capability of LMs in capturing word meanings across diachronic corpora~\cite{montanelli2023survey,tahmasebi2023computational,tahmasebi2021survey}. Current benchmarks consist of a diachronic corpus spanning two time periods $t_1$ and $t_2$ and a reference set of target words $T$ annotated with a degree of semantic change between $t_1$ and $t_2$~\cite{zamora2022lscdiscovery,kutuzov2021rushifteval,schlechtweg2020semeval,basile2020diacr}.\footnote{\citealp{kutuzov2021rushifteval} introduced a benchmark encompassing two time intervals. However, these intervals have been treated independently, leading to their consideration as two distinct sub-benchmarks over a single time interval.} The main goal is to rank the target words in $T$ according to their degree of semantic change. 

Currently, contextualized embeddings represents the state-of-the-art solution for addressing LSC. Approaches to LSC relying on contextualized embeddings are typically distinguished into two main categories: \textit{form-based} approaches and \textit{sense-based} approaches~\cite{montanelli2023survey}. The former captures semantic change by solely relying on similarities among raw embeddings without depending on sense disambiguation and representation. Given a word $w$, a common strategy involves averaging all the embeddings of $w$ from $t_1$ and all the embeddings of $w$ from $t_2$, and modeling the change as the cosine similarity of the average representations \textbf{(PRT)}~\cite{martinc2020leveraging}. The latter generally use clustering algorithm like Affinity Propagation to identify senses and subsequently model the change as divergence of cluster distributions \textbf{(JSD)}~\cite{martinc2020capturing}. 

However, when form-based approaches are employed, interpreting the detected change is not supported as they do not model each individual sense of a word. In contrast, when sense-based approaches are employed, they typically represent clusters of word usages rather than word meanings~\cite{kutuzov2022contextualized}. As a result, although a new powerful LMs has recently been introduced for modeling the semantics of words~\cite{cassotti2023xl}, the \textit{interpretation} of which meaning of a word has changed and in which way,
remains an open challenge.

In this regard, our work is related to the novel substitute-based approaches to LSC, which interpret word meaning by generating substitutes of words in context~\cite{card-2023-substitution,kudisov2022black,arefyev2020bos}. On one hand, word substitutes represents relevant keywords to aid the interpretation of senses. On the other hand, the generation process can only provide substitutes according to training data, and as we show in Section~\ref{sec:lexsubst}, LMs lack the knowledge to adapt to semantic changes. %Francesco: I would refrain to emphasize our section (If I am not wrong substitutes works better than replacements...)

To this end, we propose a novel \textit{interpretable} approach based on  
a pre-defined set of lexical \textit{replacements} rather than generated \textit{substitutions}. 

\section{Methodology}\label{sec:methodology}
In our experiments, we leverage a replacement schema to investigate the tension experienced by pre-trained LMs due to semantic change. This involves analyzing the variations in embedding representations when a target replacement is introduced. For instance, by replacing a target like \textit{cat} with a replacement like \textit{chair} in a specific context like:
\begin{center}
    The $\operatorname*{\textit{cat}}\limits_{\text{target}} \leftarrow \operatorname*{\textit{chair}}\limits_{\text{replacement}}$ was purring loudly .
\end{center} %Given our primary focus on words and their replacements when these words are split into multiple sub-words by the model, we calculate the average embeddings of the corresponding sub-words. This approach ensures the preservation of the same number of tokens in the original and artificial sentences and enables accurate distance calculations.

\subsection{The replacement schema}\label{sec:replacement-schema}
We use WordNet to generate different classes of replacements for a specific word~\cite{Fellbaum1998}, which correspond to a varying degree of plausibility (i.e. suitability of a specific replacement) between the target word and its replacement. Thus, we hypothesize that each class is associated with a different impact on contextualization. Each class of replacements also has diachronic relevance, as the synchronic, semantic relation can be considered to have a parallel in semantic change~\cite{desá2024survey,wegmann2020detecting}. %CAMERA READY: PUT IN A LINK TO Luigi's TYPE-PAPER submitted to ACL
To ensure accurate linguistic replacements, we maintain part of speech (PoS) agreement with the target words; e.g., \textit{nouns} are replaced with \textit{nouns}. % and so forth. %Examples are given in the form (target $\leftarrow$ replacement) in Table~\ref{tab:lex-repl}. %remove comment in CAMERA READY

\begin{compactitem}
\vspace{5pt}
\item \textbf{synonyms} (e.g. \textit{sadness}~$\leftarrow$~\textit{unhappiness}) are used to evaluate the stability in contextualization; that is, we hypothesize similar embeddings between target and replacement words. Indeed, synonyms are considered equally likely alternatives in LM's pre-trained knowledge. On the diachronic level, they emulate the absence of any semantic change of the replacement word;
\vspace{5pt}
\item \textbf{antonyms} (e.g. \textit{hot}~$\leftarrow$~\textit{cold}) are used to evaluate a light change in contextualization; that is, we hypothesize slightly less similar embeddings between target and replacement words. Indeed, antonyms are sometimes equally plausible alternatives, for example: \say{I \textit{love/hate} you}.  Other times they are likely to surprise the model. For example: \say{I burned my tongue because the coffee was too \textit{hot/cold}}. On the diachronic level, they emulate  a contronym change. A contronym change occurs when a word's new meaning is the opposite of its original meaning (e.g. \textit{sanction} in English) of the replacement word;
\vspace{5pt}
\item \textbf{hypernyms} (e.g. \textit{animal}~$\leftarrow$~\textit{bird}) are used similarly to \texttt{antonyms}. However, on the diachronic level, they emulate a broadening semantic change of the replacement word;
\vspace{5pt}
\item \textbf{random} words (e.g. \textit{sadness}$~\leftarrow$~\textit{eld}) are used to evaluate a change in contextualization. If LMs place high importance on the context, then the replacement should receive a similar representation to the target word. Otherwise, if LMs heavily rely on its pre-trained knowledge, the replacement will exhibit dissimilarity to the target word despite the identical context, as well as dissimilarity to the typical replacement representations. On the diachronic level, \texttt{random} emulates the presence of strong semantic change of the replacement word, that is, the emergence of a homonymic sense.
\end{compactitem}

\subsection{Data}\label{sec:data}
To avoid introducing noise into our experiments resulting from the conflation of senses, we replace words with contextually appropriate replacements based on the intended sense of the word within a specific sentence (e.g, \textit{stone} and \textit{music} for \textit{sitting on a rock} and \textit{listening to rock}, respectively). We therefore leverage the SemCor dataset~\cite{MillerLTB93}, still the largest and most commonly used sense-annotated corpus for English. To select candidate replacements,  we consider different PoS tags, namely \textit{verbs}, \textit{nouns}, \textit{adjectives} and \textit{adverbs}, and semantic classes, namely \textit{synonyms}, \textit{hypernyms} and \textit{antonyms}. We randomly sample a set of synsets for each PoS tag occurring in SemCor, and for a specific synset, we extract a subset of contexts (i.e., sentences) where a word is annotated with that synset. We sample a maximum of 10 sentences per synset to prevent oversampling of high-frequency synsets. We control for the position of the replaced target has in the sentence, and the length of the sentence, to confirm that these aspects will not bias our experiments differently across PoS. %the PoS classes are balanced in this aspect.
For each sentence, we generate the \textit{synonym} and  \textit{antonym} replacements for all PoS, and \textit{hypernym} replacements only for nouns and verbs because WordNet lacks hypernym information for other PoS (see  Table~\ref{tab:semcor_stats}). 

%PIERLUIGI: why are there very few adverb compared to other PoS?
\begin{table}[!ht]
\resizebox{\columnwidth}{!}{%
\begin{tabular}{|c|c|c|c|}
\hline
\textbf{PoS} & \multicolumn{1}{c|}{\textbf{N. target words}} & \multicolumn{1}{c|}{\textbf{\begin{tabular}[c]{@{}c@{}}Avg. N. \\ of sampled senteces\\ per target word\end{tabular}}} & \multicolumn{1}{c|}{\textbf{N. examples}} \\ \hline
\textit{noun}         & 360                                           & 3.55                                                                                                          & 1277                                      \\ 
\textit{verb}         & 433                                           & 3.45                                                                                                          & 1494                                      \\ 
\textit{adjective}    & 393                                           & 3.39                                                                                                          & 1334                                      \\ 
\textit{adverb}       & 158                                           & 3.46       & 546 \\ \hline
\end{tabular}%
}
\caption{Data statistics over  PoS, sampled from SemCor.}
\label{tab:semcor_stats}
\end{table}

\paragraph{Experimental setup} %Nina: TODO to rephrase this.
We begin by studying the tension that occurs  as a consequence of replacement %within the contextualization of words by 
focusing on the word contextualization in Section~\ref{sec:contextualization}. Next, we the use of replacements as a proxy for semantic change in Section~\ref{sec:lexical-semantic-change}. In our experiments, we use monolingual BERT\footnote{\textit{bert-base-uncased}}, mBERT\footnote{\textit{bert-base-multilingual-cased}}, and XLM-R\footnote{\textit{xlm-roberta-base}}. Our code and data are available at \url{https://github.com/ChangeIsKey/asc-lr/}. % upon acceptance. 

 \begin{figure*}[t!]
    \centering  
    \includegraphics[width=.90\textwidth]{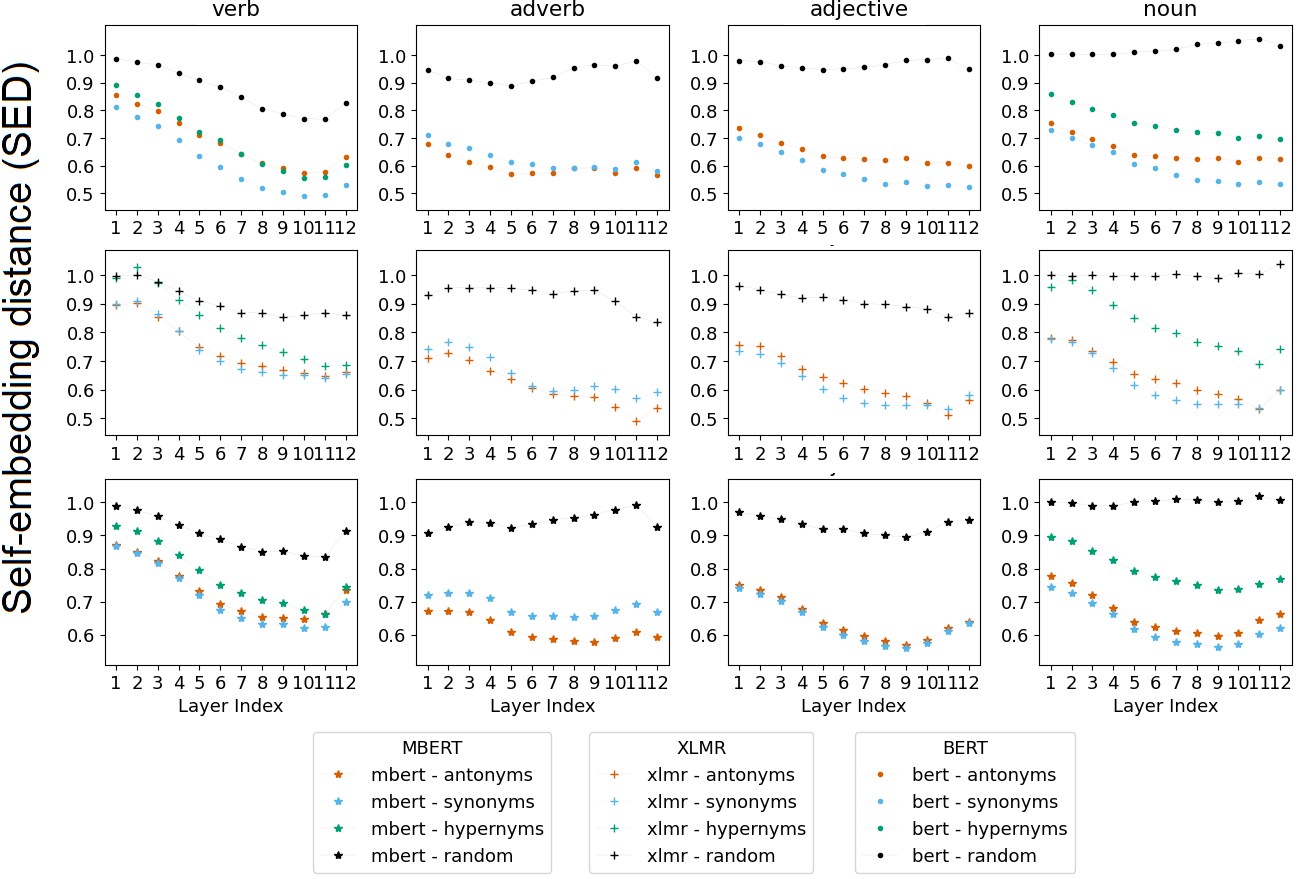} 
    \caption{Average SED over layers.}
    \label{fig:self-embedding-distance}
\end{figure*} 

\section{Tension caused by semantic change}\label{sec:contextualization} %Word contextualization
We analyze the tension experienced by LMs by comparing the embedding of a target word $w$ in the original sentence $c$ to the  embedding of the replacement word $r$ in the same sentence $c$. To perform this comparison, we rely on the cosine distance between the embeddings of $w$ and $r$. We refer to this %distance 
as the \textit{self-embedding distance} (SED). 

%, i.e.,

%\begin{center}
%\resizebox{0.8\columnwidth}{!}{%
%$\begin{matrix}\label{mat:self-emb-dist}
%c^{-n}  & ... & c^{-1} & w & c^{+1}  & ... & c^{+m}\\ 
% &   & \ & \updownarrow &   &   & \\
%c^{-n}  & ... & c^{-1} & r & c^{+1}  & ... & c^{+m}\\
%\end{matrix}
%$
%}
%\end{center}
%\noindent \newline
%where $c^{-n}, ..., c^{-1}$ and $c^{+1}, ..., c^{+m}$  denotes the embeddings of $n$ and $m$ neighbouring context words to the left and right of the target word $w$, respectively.
%\footnote{Given our primary focus on words and their replacements when 
Concretely, if $w$ and $r$ are split into multiple sub-words by the model, we calculate the average embeddings of the corresponding sub-words. This approach ensures the preservation of the same number of tokens in the original and synthetic sentences and enables accurate distance calculations.

The less plausible the relationship between the context $c$ and the replacement word $r$ for LMs, the higher the SED, leading them to rely on the pre-trained knowledge of $r$ to contextualize $r$ in context.
When there is a large mismatch between the meanings of the replacement word $r$ and the context $c$, as is the case with the random replacement, then the SED is the highest.   

\subsection{Self-embedding distance}
\label{sec:sedsection}
For each pair of original and synthetic sentences, we computed SED across each layer. %Francesco: here we can reference the GradedLSC paper and motivate why each layer.
We then analyzed the average SED for each class of replacement and PoS across the layers of LMs. It is know that contextualized embeddings experience an anisotropic nature, that is, the embeddings occupy a increasingly narrow cone within the vector space~\cite{ethayarajh2019contextual}. This means that embeddings, and thus SED scores across layers, are not comparable. To address this issue and thus compare SED  both across layers and PoS, we use a layer-specific normalization factor. 

Specifically, for normalization, we randomly sampled an additional set of 3864 sentences independent from the sets in Table~\ref{tab:semcor_stats}. For each sentence, we randomly choose a target word and replace it with a \texttt{random} replacement regardless of the PoS agreement. Then, for each layer, we computed the average SED over this set of replacements. We use the resulting SED scores as a normalization factor for each layer that represents an upper bound approximation. 
Thus, for each layer, the same normalization factor is used across all PoS and  semantic class of replacement.  This way, the normalization cannot influence the discrepancies among different classes for a specific layer but serves to make the scores in different layers somewhat comparable.\footnote{We have tested with different normalization factors -- e.g., replacing a word with a special token (``\texttt{[REPL]}'') outside the LMs vocabulary -- and found that the conclusions remain.}

Like~\newcite{ethayarajh2019contextual}, we observe that the contextualization increases across layers as the SED decreases, the context thus has a larger effect in determining the representation of a word in the higher layers. 
For \texttt{adverbs},  \texttt{adjectives} and \texttt{nouns}  the synonym and antonym classes are associated with an SED of around 0.6--0.8 in the first layer. The SED then decreases to between 0.5--0.6. For \texttt{adverb} the synonym and antonym class remain similar also in the later layers, while for \texttt{adjectives} and \texttt{nouns} we find that the synonyms have lower SED than do antonyms. For \texttt{nouns}, the hypernym class has consistently higher SED than synonyms and antonyms, despite being a more general concept where the subconcept of the target word should be contained (e.g., \textit{fruit} as a hypernym of \textit{banana}).  This aligns with the recent findings of~\newcite{hanna2021analyzing}, suggesting that BERT's understanding of noun hypernyms is limited.  

The SED score for \texttt{random} is fairly stable across all layers, meaning that when a word gains  a completely novel sense, LMs fall short in contextualizing beyond the pre-trained knowledge it has of the word. That is, the representation of the random word does not mimic the representation of the target word that it replaces. The context thus has little or no effect in determining the representation of the replacement word. 

%; that is, adverbs are more contextualized than other PoS and less pre-trained knowledge is used for their representation. This is most likely because adverbs in English are more context-dependent than the other PoS. Indeed, they can modify verbs, adjectives, adverbs, and entire sentences, in contrast to adjectives, which may only modify nouns and pronouns, and to verbs and nouns, which constitute essential components of sentences. This finding is in line with the work of~\newcite{lorge2023wacky}, which shows \textit{weak differentiation amongst the semantic classes of adverbs}. 

For \texttt{verbs}, we note a higher SED for antonyms and synonyms in comparison to other PoS, comparable to the \texttt{noun} hypernyms, starting around 0.9. However, they all drop to 0.6--0.7 by the last layers. Additionally, there is a narrower gap between the SED for the random class and those for antonyms, synonyms, hypernyms. These observations suggest that, in the earlier layers, the contextualization of verbs is less pronounced for \texttt{verbs} and that the model relies more on pre-trained knowledge. %As a result, the embeddings of verbs exhibit greater similarity to the embeddings of random words in context.

%As for \texttt{adjective} and \texttt{noun}, we note that they exhibit similar contextualization across layers. Additionally, for \texttt{noun}, \texttt{hypernyms} are less similar to the target than \texttt{antonyms} and \texttt{synonyms}.

All in all, our results suggest that models exhibit varying tension for different PoS, and for different linguistic relationships between the target and the replacement word. Conversely, we interpret these findings in the following way: there is a low degree of contextualization, and thus a high degree of tension, when there is no relationship between the word and its replacement. %We further explore this effects in the Appendix for other LMs.

\section{Semantic change}\label{sec:lexical-semantic-change}

We argue that our findings in Section~\ref{sec:contextualization} regarding the tension between a word and its context has important implications when pre-trained LMs are used for modeling semantic change as we will show in this section. 

\subsection{LCS through synthetic dataset}
\textit{Form-based} approaches can still detect this semantic change to a certain degree (as an estimate of model confusion), despite using contextualized word embeddings that are not correctly capturing a word's meaning in a novel context.
However, \textit{sense-based} approaches fall short in accurately detecting the same change. This is because \textit{sense-based} approaches require modeling meanings outside the model's pre-trained knowledge before detecting the change. Since these meanings cannot be adequately modeled when semantic change has occurred, the performance of \textit{sense-based} approaches is reduced compared to that of \textit{form-based} approaches.
%: 1) \textit{form-based} can accurately recognise semantic changes by identifying low-contextualized word occurrences (see \texttt{random} in Fig~\ref{fig:self-embedding-distance}) that the model tends to position far in the space from other embeddings of the same word; 2) conversely, \textit{sense-based} approaches fall short in accurately recognising semantic changes as they require modeling meanings outside the model pre-trained knowledge as clusters of similar word embeddings. However, these clusters cannot be properly modeled, as the embeddings for new meanings outside the pre-trained knowledge suffer from a low degree of contextualization. This implies that the semantic change cannot be recognised by the emergence of a new cluster, thus reducing performance on LSC.

We further tested these implications in the LSC task by comparing PRT (based on \textit{averaging} contextualized embeddings) and JSD (based on \textit{clustering}  contextualized embeddings) on an artificial diachronic corpus spanning two time periods (see details in Appendix~\ref{app:synthetic-LSC}). Essentially, we introduced random replacements in $C_2$ with varying probabilities to emulate different degrees of change for a set of 46 target words. Subsequently, we compared the Spearman Correlation between the scores obtained with PRT and JSD with the artificially graded score of emulated semantic change. Results using BERT are presented in Figure~\ref{fig:random-LSC} (see Appendix~\ref{app:synthetic-LSC} for additional results). Our hypothesis is that while PRT can predict changes to a fairly high degree, JSD falls short because it can only correctly model the meanings that BERT is already aware of.

As shown in the figure,  using PRT, we can  model artificial semantic changes already from layer 3. This is not the case for JSD, where we observe statistically significant correlations for only a few layers. However, the significance of performance for JSD is an artifact of BERT embeddings and does not authentically represent the simulated change. We verify this by examining the modeled clusters. While, in general, the number of clusters of AP is large~\cite{periti2022done,martinc2020capturing}, representing \textit{sense nodules}\footnote{Lumps of meaning with greater stability under contextual changes~\cite{Cruse2000-CRUAOT-2}} rather than word meanings~\cite{kutuzov2022contextualized}, we find that the injected confusion in the model due to the \texttt{random} replacements results in a very low number of clusters (typically 2, maximum of 4). We report similar results in Appendix for other languages (i.e. German, Swedish, Spanish)%TODO

\begin{figure}
    \centering
    \includegraphics[width=\columnwidth]{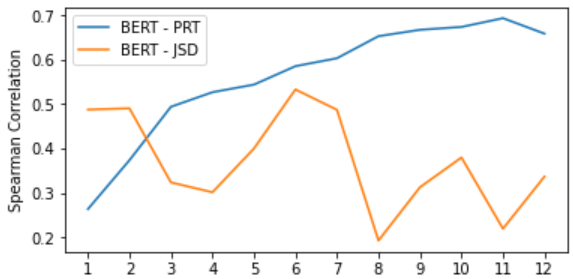}
    \caption{Spearman Correlation over layers for artificial semantic change.}
    \label{fig:random-LSC}
\end{figure}

\subsection{LSC through replacements}\label{sec:LSCreplacements}
As a further contribution of this paper, we propose a novel supervised approach to Graded Change Detection building upon the replacement schema. Our approach leverages a curated set of word replacements from WordNet and Wiktionary. 

We denote  $T=\{w_1,w_2,...,w_N\}$ as the set of target words. For each target word, we extract a set of possible replacements $\rho(w_i) = \{r_1,r_2,...,r_M\}$, resulting in $N \cdot M$ replacement pairs. The set of replacements is obtained by considering the lemmas of synonyms and hypernyms associated with the target word $w_i$ in WordNet and words extracted from the Wiktionary page corresponding to the target word. For each target word $w_i$, we sample up to 200 sentences from each period that remain stable regardless of the replacement word $r_j$. %, i.e. $S_{t_1}$ and  $S_{t_2}$. %Then, we create as far synthatic sub-corpora S_t_1 and S_t_2 as the number of considered replacements.
For each replacement pair ($w_i,r_j$), we denote the set of sentences  for a time period $t\in\{1,2\}$ as $S^t(w_i,r_j)$.%\footnote{The set of sentences in $S^t(w_i,r_j)$ are fixed for each target word $w_i$, only the replacement word $r_j$ changes.} 

%%%%%%%%%%%%%5555
For each sentence $s\in S^t(w_i,r_j)$ we measure the self-embedding distance $sed(s)$of the target and replacement word. 
%the average sentence distance is $awd (s) = \frac{1}{n+m}\sum_{i=-n}^{ m}d^i$, where $d^i$ is the self-embedding distance of the i-th word in $s$ (Section \ref{sec:repldist}). 
%%%%%%%%%%%%%%%%
%
The average self-embedding distance of a target-replacement pair is defined as \[awd^t(w_i,r_j) = \frac{1}{|S^t(w_i,r_j)|}\sum_{s \in S^t(w_i,r_j)} sed(s)\]

The absolute difference in $awd$ over time is denoted 
TD$(w_i,r_j)$. % = \left | \text{awd}^2(w_i,r_j) - \text{awd}^1(w_i,r_j) \right | \]
Finally, we rank the replacements $\rho(w_i)$ according to their degree of time difference:
\begin{small}
\[R(\rho(w_i)) = \{r_1,r_2,...,r_M | \ \text{TD}(w_i,r_{i+1},)\leq \text{TD}(w_i,r_{i})\}\]
\end{small}
and we compute a semantic change score $lsc_{w}$ as the average TD considering the top $k$ replacements: 
\[lsc_w = \frac{1}{k}\sum_{r \in R(\rho(w_i))_k} \text{TD}(w_i,r)\] %where $R(\rho(w))_k = {v_1,v_2,...,v_k}$.

\begin{figure}
    \centering
    \includegraphics[width=\columnwidth]{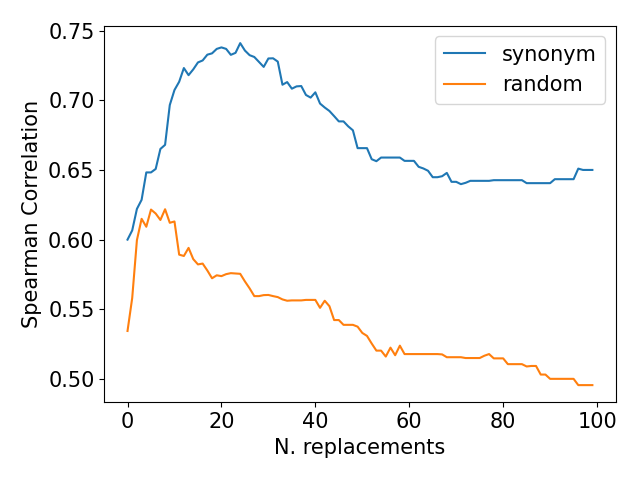}
    \caption{Top-k replacement vs Spearman Correlation.}
    \label{fig:spearman-top-k-replacement}
\end{figure}

We evaluate our approach on the SemEval-2020 Task 1, Subtask 2 dataset for English. We compute the Spearman Correlation between the graded score reported in the gold truth and the $lsc$ scores. Figure~\ref{fig:spearman-top-k-replacement} reports the correlation computed for different values of $k$. The highest correlation of 0.741 is achieved when considering the first 22 replacements, while the lowest correlation of 0.600 is obtained using only the first replacement (see Table~\ref{tab:eng_lsc_results}). Interestingly, the minimum correlation obtained using the replacements is competitive with SOTA results. Moreover, on average, the correlation is higher than the SOTA model's performance. The replacements are reported in Table \ref{ref:eng_changes}. 

%We used the linguistically-aware replacement strategy in the LSC task to assess and quantify the semantic changes undergone by words over time. 
By replacing the target words with different semantically related words, we generate contextual variations that enable the detection of semantic shifts. In the case of words like \textit{record} (attainment, track record $\longrightarrow$ evidence, document) and \textit{land} (real estate, real property $\longrightarrow$ realm, country) that have undergone semantic change through narrowing and generalisation, respectively, linguistically aware replacements can provide valuable insights.  The replacement process generates a list of replacements that can be used as labels for the types of semantic change observed. By associating each replacement with a specific semantic category or change type, it becomes possible to analyze and quantify the semantic shifts experienced by words over time. The method can also be combined with a priori clustering to get changes specific to a sense. %Francesco: check this last sentence, is it true?

\begin{table}[!ht]
\centering

\resizebox{\columnwidth}{!}{%
\begin{tabular}{ccc}
 & \textbf{Model} & \textbf{Spearman Correlation} \\ \cline{2-3} 
 & \citeauthor{rosin2022temporal} & 0.629 \\
 & \citeauthor{kutuzov2020uio} & 0.605 \\
 &  \citeauthor{laicher2021explaining} & 0.571 \\
 &  \citeauthor{periti2022done} & 0.512 \\
  &  \citeauthor{cassotti2023xl} (XL-LEXEME) & 0.757 \\ \hline
\multirow{3}{*}{\textbf{\begin{tabular}[c]{@{}c@{}}Synonym \\ Replacement\end{tabular}}} & Replacement Min. Corr. & 0.600 \\
 & Replacement Max. Corr. & 0.741 \\
 & Replacement Avg. Corr. & 0.674 \\
\hline
\multirow{3}{*}{\textbf{\begin{tabular}[c]{@{}c@{}}Random \\ Replacement\end{tabular}}} & Replacement Min. Corr. & 0.495 \\
 & Replacement Max. Corr. & 0.622 \\
 & Replacement Avg. Corr. & 0.542
\end{tabular}%
}
\caption{Spearman Correlation on SemEval-2020 Task 1 (Eng)}
\label{tab:eng_lsc_results}
\end{table}

\paragraph{Random replacements}
Here, we focus on the results using randomly selected words with the same PoS as the target word, i.e. \texttt{random} replacement as introduced in Section~\ref{sec:methodology}. This approach generates a list of replacement words contextually unrelated to the target word. Some interesting patterns emerge when these results are compared with those obtained using synonym replacement. In the case of semantic change detection, the use of synonyms can provide more contextually relevant replacements, as they share semantic relationships with the target word. However, using random  replacements can still yield reasonable results, as evidenced by an average correlation of 0.542. These results is in line with the finding of Section~\ref{sec:lexical-semantic-change}.

In this approach, although random replacements tend to perform worse than synonym replacements, they have one distinct advantage: they do not rely on external lexical resources and are thus suitable for unsupervised scenarios. While synonym replacements can improve contextualization and semantic relevance, they are not always readily available or reliable for languages with limited linguistic resources. In such cases, random  replacements can still provide reasonable results and serve as a practical, resource-efficient approach for tasks where synonym information is scarce or unavailable.

In Section \ref{sec:sedsection}, when using SemCor, we effectively account for the nuances of different word senses, thereby improving the contextualization and semantic relevance of synonym replacements. This approach is more targeted as synonyms are selected based on their association with a particular sense, leading to higher quality contextualization in the context of that sense. As a result, synonym replacements are more finely tuned to the specific meaning of the target word, reducing noise and improving correlation with semantic change labels.

%The lower correlations observed with random replacements indicate that contextualization effects vary significantly when unrelated words are introduced into the context. This emphasizes the crucial role of context in the performance of language models when they have prior knowledge of input word meanings. However, it should not be assumed that contextualization is equally effective when modeling new meanings outside the scope of the model's pre-trained knowledge.

\subsection{LSC through substitutions}\label{sec:lexsubst}
Finally, we assess the use of lexical substitutes generated by LMs for LSC. By asking LMs' to generate substitutions, we probe them for their information about the target word given the context. Similar to \citet{card-2023-substitution,arefyev2020bos}, we use monolingual BERT. We additionally compared the use of a larger, generative model such as LLaMa 2 7B~\cite{touvron2023LLaMa}\footnote{\textit{meta-llama/Llama-2-7b-hf}}.  

%These models have varying numbers of parameters, and the size of the training data differs significantly. %We hypothesize that the number of parameters as well as the both the type and amount of training data  influence their ability to understand the meaning a word takes in a specific context.

For BERT, we use the masking strategy, meaning that we mask a target word with the special token and generate possible substitutes. For LLaMa 2, we fine-tune the model to enable it to predict the target word. Specifically, we fine-tune LLaMa 2 by inputting the original sentence, adding two asterisks at the beginning and end of the target word.  Following the sentence we provide the list of substitutes found in ALaSCA \cite{lacerraetal:2021}, the largest existing dataset for lexical substitution:\\

\noindent {\small During the siege, George Robertson had appointed Shuja ul-Mulk, who was a \textbf{**}bright\textbf{**} boy only 12 years old and the youngest surviving son of Aman ul-Mulk, as the ruler of chitral. \textit{\textbf{|answer|} intelligent \textbf{|s|} clever \textbf{|s|} smart \textbf{|end|}}}\\

\noindent where \textbf{|answer|}, \textbf{|s|}, and \textbf{|end|} are added as special tokens in the model. For efficiency reasons, we train the model using the QLoRA paradigm\footnote{Quantization and Low-Ranking Adaptation \cite{dettmers2023qlora}}. We fine-tuned for one epoch using a learning rate of 2e-4, and set the LoRA configuration with a rank of 8 and an alpha of 16.

The data used for the evaluations is the same  in Section \ref{sec:LSCreplacements}. In Table \ref{table:sub-example} we report an example of the generated substitutions. To calculate the degree of semantic change, we consider all uses of a word in time periods  $t_1$ and $t_2$. We consider the substitutes generated for each usage and calculate the distance between all possible pairs of uses between $t_1$ and $t_2$. To calculate the distance, we use the Jaccard Distance between the sets of generated substitutes. Lastly, the Jaccard distances are averaged, and we use the average as a score for LSC. In Table \ref{tab:substitution_results} we show the result on the SemEval 2020 Task 1 - Subtask 2 (other comparable results in Table \ref{tab:eng_lsc_results}). Our results for BERT are somewhat comparable with SOTA results, while being lower to those obtained through lexical replacements, likely because the replacements are of higher quality when  found using WordNet, while the substitutions are generated by the model with its limited knowledge of the context. In contrast, our results for LLaMa 2 are even higher than the results obtained with lexical replacements achieving comparable performance to the one obtained with XL-LEXEME. We attributed this higher performance to the fact that both LLaMa and XL-LEXEME have been fine-tuned on generating lexical substitutes and WiC task, respectively which, rather than using all of the model's pre-trained knowledge, forces the model to focus on the semantic aspect specifically.

%We clearly see the gap between the smaller LMs (also the BERT-based models used in \cite{arefyev2020bos} and \cite{card-2023-substitution}) and LLaMa 2. The performance of the latter resembles that of the task-specific fine-tuned model XL-LEXEME (i.e., 0.757), indicating that either the size of the pretrained data matters, or the model has, directly or indirectly, been fine-tuned on the WiC task similar to XL-LEXEME. 

\begin{table}
\small
    \centering
    \begin{tabular}{cc}
         \textbf{Model}& \textbf{Spearman Correlation}\\
         \hline
         \citet{arefyev2020bos} & 0.299\\
         \citet{card-2023-substitution} & 0.547\\\hline
         \textbf{LLaMa 2 7B}& \textbf{0.731}\\
         \textit{BERT} & \textit{0.450}\\
         %\textit{XLM-R}& \textit{0.380}\\
 \bottomrule
    \end{tabular}
    \caption{Spearman Corr. on SemEval-2020 Task 1 (EN)}
    \label{tab:substitution_results}
\end{table}

\section{Discussions and Conclusions}
In this paper, we study semantic change using lexical replacement. From the point of view of the replaced word, a semantic change takes place as the  word gains contexts which it has not encountered previously. When the replacement is closely related to the target word, for example by synonymy, the novelty of the context for the replacement word should be low. However, novelty will increase as the relation between the target and replacement becomes more distant. We are assuming that the replacements based on synchronic relations will offer insights into semantic change diachronically.

To test this hypothesis, we used self-embedding distance (SED) when the context stays the same, using all layers of BERT, mBERT and XLM-R across four part of speech. Not surprising, we found that the self-embedding distance  is smallest for synonym replacements and highest for the random replacements. And like~\citet{ethayarajh2019contextual}, we found that more contextualization happens across the last layers. For the different models, we also find slightly different behaviors. However, consistently, adverbs and adjective have lower SED scores than  verbs and nouns. We show that hypernymy is a more distant relation for LMs than antonymy and synonymy

%Next, we used the replacement schema to model lexical semantic change. When introducing artificial semantic change using random replacements, we found that while comparing averaged contextualized embeddings over time work decently well, clustering does not. We hypothesis that this stems from the models inability to  model random replacements. %the previous part I’m to sure about

We then employ replacements for measuring the degree of semantic change. For this, we generate synonym replacements using WordNet, for each word in the English portion of the SemEval-2020 Task 1 benchmark. We assume that if a word has not experienced semantic change, the SED between the replacements and the target word are similar across time. If however, a word has experienced semantic change resulting in context changes, SED scores will be different over time as the replacements will be more distantly related to the contexts. This method offers a novel \textit{interpretable} semantic change detection. Finally, we ask the LMs themselves to generate substitutions for a target word in the English SemEval data. %We compare these using a version of the average pairwise distance (APD) method replacing the cosine similarity with  Jaccard similarity between the sets of replacements. 

\section{Limitations}
A potential limitation of our study lies in the use of the replacement schema in conjunction with lexical replacements generated from WordNet: inherent limitations of WordNet, such as potential gaps, inaccuracies, or ambiguities in the semantic relationships may influence our analysis. WordNet also limits the data sources from which we can draw sentences, since we need a corpus with sense annotations corresponding to a lexicon. 

Furthermore, in our experiment, the lexical replacement process involves replacing a \textit{word} occurrence in the original sentence with a related \textit{lemma} extracted from WordNet. As a result, providing the model with synthetic sentences containing the lemma instead of the inflected word may influence the generation of word embeddings and the contextualization of every word in the sentences. However, we assume that this limitation equally affects every class we consider and all models. For example, while the lemma of a verb may reduce the third singular verb form, the plural forms of adjectives and nouns can also be simplified to singular lemma forms. Additionally, to mitigate these issues and ensure that all PoS are equally affected by the replacement procedure, we replaced both the target and replacement words with lemmas in the original and synthetic sentences, respectively. We did not analyze semantic change in Section 5 with respect to different PoS because there are no available LSC benchmarks with a substantial number of targets for different PoS, nor any sense-tagged benchmarks except for a small subset for German. 

Finding the correct form of a replacement requires advanced morphological analysis and carries the risk of leading to errors. For now, we therefore opted to circumvent this by replacing targets and lemmas alike. Furthermore, we would like to highlight a relevant study by~\newcite{laicher2021explaining} that delves into the influence of various linguistic variables on the use of BERT embeddings for the LSC task. This research demonstrates that by reducing the influence of orthography through lemma usage, significant enhancements in BERT's performance were observed for German and Swedish, while maintaining comparable results for English. This underscores the potential benefits of lemma-based contextualization and that linguistic features like orthography can sometimes be minimized without substantial loss of performance.

We use LLaMa 2 only for our last experiment. This stems from the difficulty to generate contexualized representation of a single word in context in LLMs. We also do not exhaustively test LLMs as this lies outside the scope of the paper, while requiring a lot more resources. Instead we use one open LLM to test the knowledge of a model when trained on significantly more data. 

Finally, in the introduction, we use the example of \textit{gay} and \textit{happy} to illustrate that word \textit{happy} is replaced in contexts of \textit{gay} for the meaning of \texttt{happiness}. We are however aware that happy gained the meaning of \texttt{happiness} several hundred years before \textit{gay} lost its sense of \texttt{happiness}, and only use the example for illustrative purposes. 

\section*{Acknowledgements}
This work has in part been funded by the project Towards Computational Lexical Semantic Change Detection supported by the Swedish Research Council (2019–2022; contract 2018-01184), and in part by the research program Change is Key! supported by Riksbankens Jubileumsfond (under reference number M21-0021). The computational resources were provided by the National Academic Infrastructure for Supercomputing in Sweden (NAISS), partially funded by the Swedish Research Council through grant agreement no. 2022-06725.

We would also like to thank the anonymous reviewers for their helpful comments, and Stefano De Pascale, David Alfter, and Dirk Geeraerts for providing valuable feedback on the preliminary draft of this work, as well as for engaging in early discussions that contributed to the development of this research.

\bibliography{bib}

\clearpage
\section*{Appendix}
\appendix

\begin{table}[!b]
\centering
\resizebox{\columnwidth}{!}{%
\begin{tabular}{ccc}
\textbf{References} & \textbf{Benchmark} & \textbf{\# targets} \\
\citealp{schlechtweg2020semeval} & DWUG-English & 46 \\
\citealp{schlechtweg2020semeval} & DWUG-German & 50 \\
\citealp{schlechtweg2020semeval} & DWUG-Swedish & 44 \\
\citealp{zamora2022lscdiscovery} & DWUG-Spanish & 100
\end{tabular}%
}
\caption{References and number of targets for each consider artificial corpus}
\label{tab:artificial-corpora}
\end{table}

\begin{figure}[!h]
 \centering
    \includegraphics[width=\columnwidth]{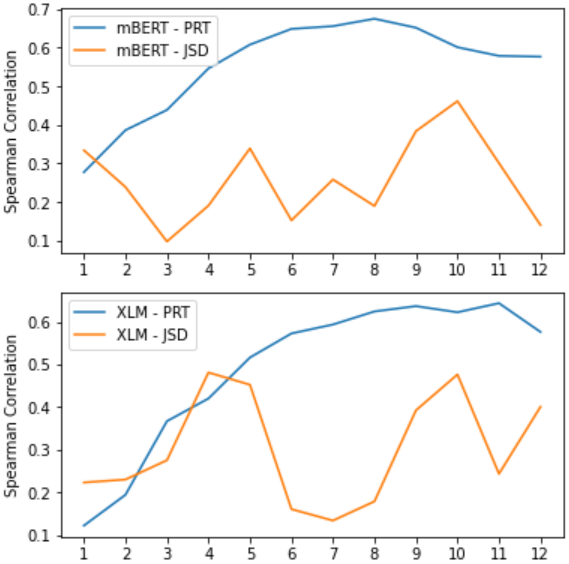}
    \captionof{figure}{PRT and JSD performance on the artificial LSC dataset}
    \label{fig:random-LSC-models}
 \end{figure}

\section{\texttt{Random} Lexical Semantic Change}\label{app:synthetic-LSC}
\subsection{Artificial diachronic corpus}
We generated an artificial diachronic corpus for LSC by utilising the SemEval and LSCDiscovery benchmakrs for LSC in DWUG format\footnote{English: \url{https://zenodo.org/records/5796878}, German: \url{https://zenodo.org/records/5796871}, Swedish: \url{https://zenodo.org/records/5090648}, Spanish: \url{https://zenodo.org/records/6433667}} (see Table \ref{tab:artificial-corpora}). Instead of incorporating data from both time periods, $T_1$ and $T_2$, we discarded information from the first time period as it is more likely to contain word meanings outside the pre-trained knowledge of the models under examination. We created two distinct artificial sub-corpora, $C_1$ and $C_2$, by randomly sampling occurrences from the data of the second time period $T_2$. The DWUG English dataset contains data for 46 target words.

For each target $t$, we considered all sentences where another target $t1$, with $t1 \neq t$, appeared as potential candidates to emulate instances of semantic change. We simulated a change instance through a \textit{random} replacement, that is by replacing $t$ in the sentence where $t1$ occurred -- i.e., $t1 \leftarrow t$. We sample a varying number of sentences and perform replacements for each target, thereby emulating a varying degree of semantic change.

\clearpage
\section{Lexical Semantic Change}

\begin{minipage}[t]{\columnwidth}
\centering
\resizebox{\columnwidth}{!}{%
\begin{tabularx}{5.5in}{|l|l|X|r|}
\hline
Word                    & Time span & (Ranked) Farthest replacements                                                                                                    & \multicolumn{1}{l|}{$lsc_w$ (k=1)} \\ \hline
\multirow{2}{*}{attack} & T1        & \textbf{physical}, degeneration, blast, crime, disease, death, condition, plane, affliction, birthday attack                                     & -0.036                      \\ \cline{2-4} 
                        & T2        & \textbf{approach}, force, onslaught, assault, exploit, challenge, commencement, aim, worth, signal                                               & 0.059                       \\ \hline
\multirow{2}{*}{bit}    & T1        & \textbf{nominative case}, accusative case, cryptography, information theory, bdsm, time,point, binary digit, sociologic, sublative              & -0.018                   \\ \cline{2-4} 
                        & T2        & \textbf{saddlery}, chard, illative case, iron, bevelled, tack, small, gun, cut, elative case                                                     & 0.067                       \\ \hline
\multirow{2}{*}{circle} & T1        & \textbf{wicca}, circumlocution, encircle, astronomy, tavern, semicircle, around, logic, go,wand                                                 & 0.002                    \\ \cline{2-4} 
                        & T2        & \textbf{pitch}, place, graduated, figure, disk, territorial, enforce, worship, line, bagginess                                                   & 0.064                      \\ \hline
\multirow{2}{*}{edge}   & T1        & \textbf{brink}, cricket, instrument, margin, polytope, side, edge computing, verge, demarcation line, demarcation                                & -0.015                      \\ \cline{2-4} 
                        & T2        & \textbf{data}, production, climax, division, superiority, organization, sharpness, graph, win, geometry                                          & 0.047                      \\ \hline
\multirow{2}{*}{graft}  & T1        & \textbf{lesion}, bribery, felony, politics, bribe, corruption, autoplasty, surgery, nautical, illicit                                            & -0.047                     \\ \cline{2-4} 
                        & T2        & \textbf{branch}, stock, tree, fruit, shoot, join, cut, graft the forked tree, stem, portion                                                      & 0.103                      \\ \hline
\multirow{2}{*}{head}   & T1        & \textbf{headland}, head word, capitulum, syntactic, pedagogue, fluid dynamics, hip hop, headway, pedagog, word                                   & 0.004                      \\ \cline{2-4} 
                        & T2        & \textbf{leader}, organs, implement, top, tail, foreland, chief, bolt, axe, forefront                                                             & 0.084                     \\ \hline
\multirow{2}{*}{land}   & T1        & \textbf{real estate}, real property, surface, property,build, physical object, Edwin Herbert Land, electronics, landing, first person           & -0.032                      \\ \cline{2-4} 
                        & T2        & \textbf{realm}, country, kingdom, province, domain, people, homeland, territory, nation, region                                                  & 0.076                      \\ \hline
\multirow{2}{*}{lass}   & T1        & \textbf{sweetheart}, girl, missy, woman, yorkshire, lassem, lasst, lassie, loss, miss                                                            & 0.014                      \\ \cline{2-4} 
                        & T2        & \textbf{fille}, dative case, jeune fille, loose, lasses, unattached, young lady, young woman, north east england, past participle                & 0.099                    \\ \hline
\multirow{2}{*}{plane}  & T1        & \textbf{airplane}, aeroplane, pt boat, heavier-than-air craft, glide , boat, lycaenidae, lift, bow, hand tool                                     & -0.197                   \\ \cline{2-4} 
                        & T2        & \textbf{geometry}, point, shape, surface, flat, degree, form, range, anatomy, smooth                                                             & 0.205                  \\ \hline
\multirow{2}{*}{player} & T1        & \textbf{media player}, idler, soul, thespian, person, individual, trifler, performer, somebody, histrion                                         & -0.065                      \\ \cline{2-4} 
                        & T2        & \textbf{contestant}, performing artist, actor, musician, musical instrument, music, gamer, theater, player piano, play the field                 & 0.042                      \\ \hline
\multirow{2}{*}{prop}   & T1        & \textbf{props}, airscrew, astronautics, actor, airplane propeller, seashell, stagecraft, stage, property, art                                    & -0.042                    \\ \cline{2-4} 
                        & T2        & \textbf{around}, rugby, imperative mood, about, singular, scrum, ignition, roughly, ballot, manually                                             & 0.088                      \\ \hline
\multirow{2}{*}{rag}    & T1        & \textbf{ragtime}, nominative case, accusative case, rag week, terminative case, inflectional, sublative, piece of material, tag, sanitary napkin & -0.049                      \\ \cline{2-4} 
                        & T2        & \textbf{clothes}, exhaustion, university, society, silk, ragged, journalism, haze, ranking, torment                                              & 0.071                     \\ \hline
\multirow{2}{*}{record} & T1        & \textbf{attainment}, track record, achievement, accomplishment, struct, number, intransitive, record book, criminal record, disc                 & -0.036                     \\ \cline{2-4} 
                        & T2        & \textbf{evidence}, document, information, audio, recollection, storage medium, memory, electronic, sound recording, data                         & 0.089                     \\ \hline
\multirow{2}{*}{stab}   & T1        & \textbf{thread}, staccato, feeling, nominative case, sheet, chord, bacterial, culture, twinge, sensation                                         & -0.046                     \\ \cline{2-4} 
                        & T2        & \textbf{wound}, tool, knife thrust, weapon, plaster, criticism, wire, pierce, thrust, try                                                        & 0.029                     \\ \hline
\multirow{2}{*}{thump}  & T1        & \textbf{clunk}, throb, clump, thud, pound, thumping, rhythmic, sound, blow, hit                                                                  & -0.036                    \\ \cline{2-4} 
                        & T2        & \textbf{muffled}, hit, blow, sound, rhythmic, thumping, pound, thud, clump, throb                                                                & 0.033                     \\ \hline
\multirow{2}{*}{tip}    & T1        & \textbf{gratuity}, first person, forty, bloke, singular, overturn, stringed instrument, unbalanced, taxi driver, sated                           & -0.031                      \\ \cline{2-4} 
                        & T2        & \textbf{brush}, tap, strike, gift, tram, flex, tumble, heap, full, hint                                                                          & 0.070                     \\ \hline
\end{tabularx}%
}
\captionof{table}{Words annotated as changed in SemEval 2020 Task 1: Binary Subtask and retrieved farthest replacements for each time span.}
\label{ref:eng_changes}
\end{minipage}

\clearpage

\begin{figure*}[!h]
 \centering
    \includegraphics[width=\textwidth]{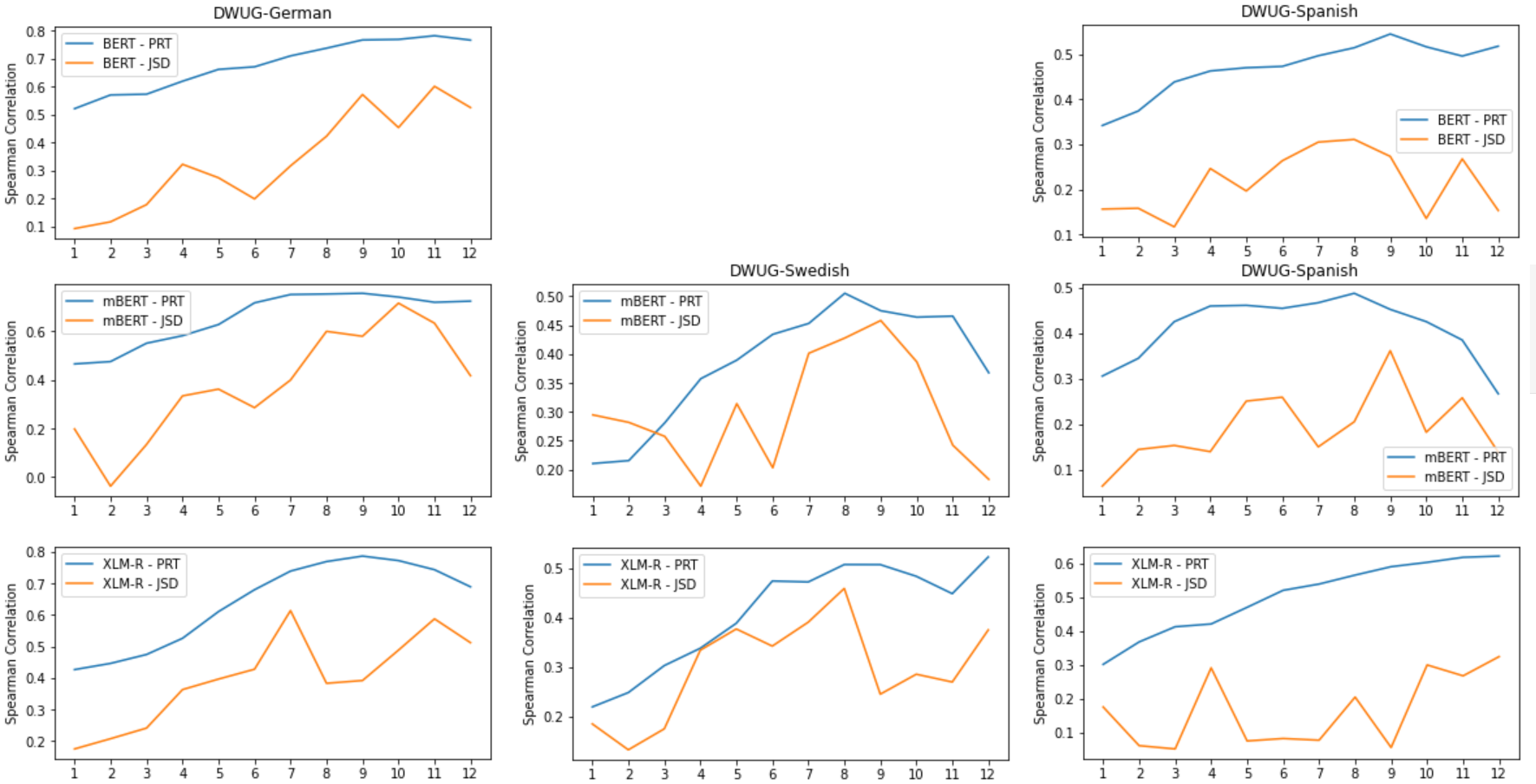}
    \captionof{figure}{PRT and JSD performance on the artificial LSC dataset}
    \label{fig:random-LSC-languages}
 \end{figure*}

\begin{table*}[]
\resizebox{\textwidth}{!}{%
\begin{tabular}{lp{6cm}p{6cm}}
 & \multicolumn{1}{c}{\textbf{T1}} & \multicolumn{1}{c}{\textbf{T2}} \\
 \hline
 & remember that it be only such line as be nearer the ground \textbf{plane} than the eye that be draw under the horizon line & as his \textbf{plane} cross north carolina and head south over the atlantic it pick up a small convoy of escort military craft that try to make radio contact but fail \\
 \hline
\textbf{BERT} & there, be, where, here, and & planes, over, out, boats, aircraft \\

%\textbf{XLM-R} & line, rather, and, more, level & planes, crew, men, vehicles, team \\

\textbf{LLaMa 2} & level,surface,flat plane,horizontal plane & aircraft,airplane,jet,plane model,propeller-driven vehicle
\end{tabular}%
}
\caption{Generated substitutions for usages of \textbf{plane} extracted by SemEval 2020 Task 1 English.}
\label{table:sub-example}
\end{table*}

\end{document}